\documentclass[10pt,twocolumn,letterpaper]{article}

\usepackage{cvpr}
\usepackage{times}
\usepackage{epsfig}
\usepackage{graphicx}
\usepackage{amsmath}
\usepackage{amssymb}
\usepackage{subcaption}
\usepackage{graphicx}
\usepackage{float}

\usepackage{graphicx}
\usepackage{caption}

\usepackage[pagebackref=true,breaklinks=true,letterpaper=true,colorlinks,bookmarks=false]{hyperref}

\cvprfinalcopy 

\def\cvprPaperID{****} 

\ifcvprfinal\fi

\newcommand\blfootnote[1]{%
  \begingroup
  \renewcommand\thefootnote{}\footnote{\tiny{#1}}%
  \addtocounter{footnote}{-1}%
  \endgroup
}

\title{Efficient Surfel Fusion Using Normalised Information Distance}

\author{Louis Gallagher and
John B. McDonald \\
Maynooth University, Department of Computer Science\\
Maynooth, Co. Kildare, Ireland\\
{\tt\small Louis.Gallagher@mu.ie}
{\tt\small johnmcd@cs.nuim.ie}\\
}

\begin{document}

\twocolumn[{%
\renewcommand\twocolumn[1][]{#1}%
\maketitle
\begin{center}
  \centering
  \begin{tabular}{ccc}

  \includegraphics[scale = 0.2]{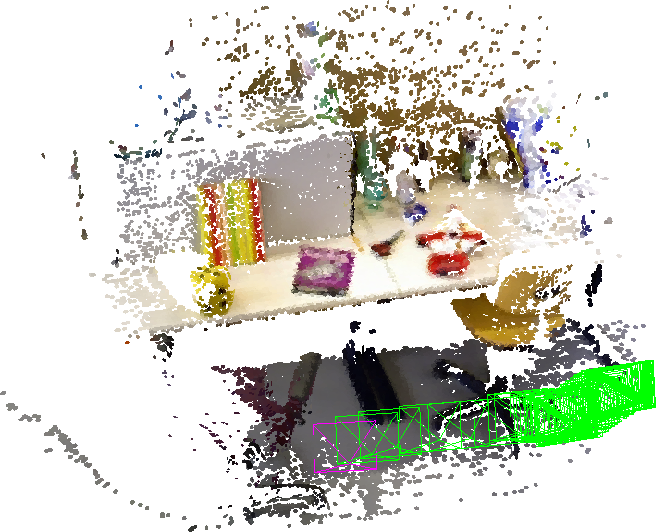}\label{nid_ex_1} & 
  \includegraphics[scale = 0.2]{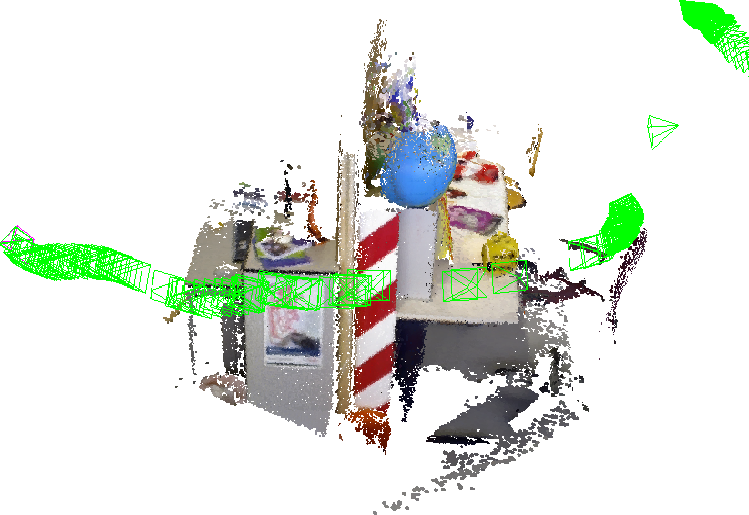}\label{nid_ex_3} &
  \includegraphics[scale = 0.2]{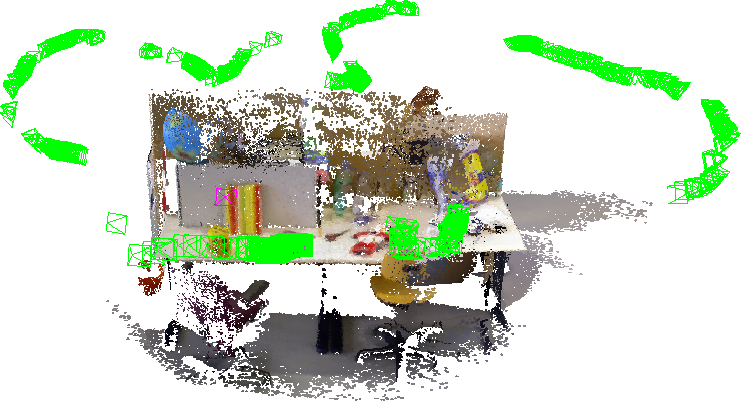}\\ 
    (i)   & (ii) & (iii)\\
  \end{tabular}
  \captionof{figure}{A handheld camera explores a scene taken from the TUM RGB-D dataset \cite{freiburg}. Green frustums represent the pose of the camera for fused frames. (i) Initially the NID between frames predicted from the incomplete model and live frames is quite high so all frames are fused. (ii) The camera continues exploring but selectively decides not to fuse frames that the NID deems are well explained by the model. (iii) The camera closes a loop and transitions fluidly to purely tracking against reactivated portions of the model as there is now no need for fusion.}
  \label{nid_example}
\end{center}%
}]


\begin{abstract}
  \vspace{-1em}
   We present a new technique that achieves a significant reduction in the quantity of measurements required for a fusion based dense $3D$ mapping system to converge to an accurate, de-noised surface reconstruction. This is achieved through the use of a \textit{Normalised Information Distance} metric, that computes the novelty of the information contained in each incoming frame with respect to the reconstruction, and avoids fusing those frames that exceed a redundancy threshold. This provides a principled approach for opitmising the trade-off between surface reconstruction accuracy and the computational cost of processing frames.
   The technique builds upon the \textit{ElasticFusion} (EF) algorithm where we report results of the technique's scalability and the accuracy of the resultant maps by applying it to both the ICL-NUIM~\cite{iclnuim} and TUM RGB-D~\cite{freiburg} datasets. 
   These results demonstrate the capabilities of the approach in performing accurate surface reconstructions whilst utilising a fraction of the frames when compared to the original EF algorithm. 
   \blfootnote{This research was supported, in part, by the IRC GOIPG scholarship scheme grant GOIPG/2016/1320 and, in part, by Science Foundation Ireland grant 16/RI/3399 and grant 13/RC/2094 to Lero - the Irish Software Research Centre (www.lero.ie).}
   \end{abstract}   
   \section{Introduction \& Background}\label{intro}
   Dense reconstruction algorithms take advantage of multiple overlapping surface measurements to estimate a denoised surface model. BundleFusion~\cite{bundlefusion}, ElasticFusion~\cite{elasticfusionjournal}, Kintinuous~\cite{kintinuous} and KinectFusion~\cite{kinectfusion} are all exemplars of this class of algorithms. Common to each of these algorithms is a focus on maximising the accuracy of both the reconstructed models and estimated $6$-DOF motion of the camera without considering the relationship between the cost of processing and its impact on model quality. In this paper we propose an general information theoretic approach to modelling this relationship and demonstrate its potential in significantly reducing the processing requirements of dense surfel fusion with limited impact on accuracy.

   Although such considerations are not typically required in the context of a high end GPU processing the output from a single camera, they become critical when applying these algorithms in both collborative settings and in compute limited scenarios. For example, recent work on the development of collaborative dense SLAM showed that increasing the number of sensors in the fusion process rapidly leads to a degradation in system performance, which drops from real-time processing rates of more than $25hz$ for $1$ and $2$ camera sessions to interactive rates of just $12hz$ for $3$ camera sessions~\cite{collabef}. In compute limited scenarios common in much of robotics and augmented reality, additional costs such as battery usage and heat dissipation highlight the need for more efficient approaches.

   Since the seminal KinectFusion algorithm was introduced there has been a profusion of methods proposed that use scalable data structures, such as hierarchical trees and hash tables, for efficiently storing the underlying TSDF volume\cite{octree, voxel_hashing}.
   Geometric simplification methods that reduce the number of parameters used to represent the map, such as incremental meshing
   have also been proposed \cite{ incremental_mesh}. These methods, however, must maintain complexity around surface discontinuities and regions of high frequency detail and require intermediate map representations from which they infer simpler models.\par
   Methods that leverage efficient algorithms for camera tracking and fusion, compact map representations and geometric simplifications for optimising the whole SLAM pipeline have also been proposed \cite{kintinuous, infinitam_efficient}. \par
   In comparison relatively little attention has been paid to increasing the efficiency of surfel-based SLAM methods. 
   In model-predictive frameworks, like EF, as map size and complexity increase so too does the cost of making map predictions. In the presence of sensor noise and drift in camera tracking simply fusing every frame can lead to spurious surfels and regions of the surface being over represented by the map.
Though the above approaches could potentially be helpful in improving the efficiency of surfel fusion it is the purpose of this work to explore a mode of keyframing that is based on quantifying the novelty of the information in incoming frames. The method we propose more optimally balances the accuracy of dense surfel models against the computational cost of dense surfel fusion. To demonstrate the effectiveness of the method we place it in the pipeline of the EF mapping system \cite{elasticfusionjournal} and report initial empirical evidence that this leads to good precision and scalability characteristics in comparison to the original EF algorithm. The contributions of this work are:~(i) A method for subsampling frames in a real-time dense reconstruction pipeline and (ii)~An empirical analysis of this method.

   \section{Approach}\label{approach}
   In model-predictive camera tracking, once the current pose of the camera, $P_t \in \mathbb{SE}_3$, has been estimated it is used to render a synthetic view of the map, $\mathcal{F}_t^p$.
   For each frame, we compute 
   the normalised information distance ($NID$) between the appearance distribution of $\mathcal{F}_t^p$ and that of the live frame, $\mathcal{F}_t^l$, 
   
   \begin{equation}
     NID(\mathcal{F}_t^p, \mathcal{F}_t^l) = \frac{H(\mathcal{F}_t^p, \mathcal{F}_t^l) - I[\mathcal{F}_t^p; \mathcal{F}_t^l]}{H(\mathcal{F}_t^p, \mathcal{F}_t^l)}
   \end{equation}
   where, $H(X,Y)$ and $I[X;Y]$ are the joint entropy\footnote{$H(X, Y) = \sum_{x,y}P(x,y)log_2P(x,y)$} and mutual information\footnote{$I[X;Y] = H(X) + H(Y) - H(X, Y)$}, respectively, of two random variables $X$ and $Y$. When the NID is high this implies that the live camera frame is not well explained by the current model estimate, and hence the system should fuse that frame. See Figure \ref{nid_example} and its caption for a detailed example and explanation for NID based sampling.\par 
   
   
   
   For each frame we compute two NID scores, one for the RGB component of the frame, $NID_{rgb}$, and one for the depth component, $NID_d$. The final NID is the weighted sum
   \begin{equation}
     NID(\mathcal{F}_t^l,\mathcal{F}_t^p) = \alpha NID(\mathcal{I}_t^l, \mathcal{I}_t^p) + (1 -\alpha)NID(\mathcal{D}_t^l, \mathcal{D}_t^p)
   \end{equation}
   where, $\alpha \in [0..1]$ denotes the relative weighting between the RGB and depth components, $\mathcal{I}$ denotes the intensity image derived from the color channels of the frame, and $\mathcal{D}$ denotes the depth component.\par
   To compute both NID scores we follow the approach outlined in \cite{Stewart_info}, except we drop the cubic spline weighting as we do not require a differentiable histogram. We compute a joint appearance intensity histogram and a joint appearance $2.5D$ depth histogram. Each of the bins in the histograms corresponds to a range of values in $\mathcal{F}_t^l$ co-ocurring with a range of values in $\mathcal{F}_t^p$.
   The histogram $\mathcal{P}$ is computed in a SIMD fashion on the GPU using CUDA atomic instructions then the marginal entropies and, ultimately, the NID is derived on the CPU by marginalising over the histogram's rows and columns.\par
   
   
   
   \subsection{Accounting for Time}
   \begin{figure*}[t]
    \label{icl_qualitative}
    \centering
  \begin{subfigure}[t]{0.30\linewidth}\centering\includegraphics[width=4.5cm, height=3.5cm]{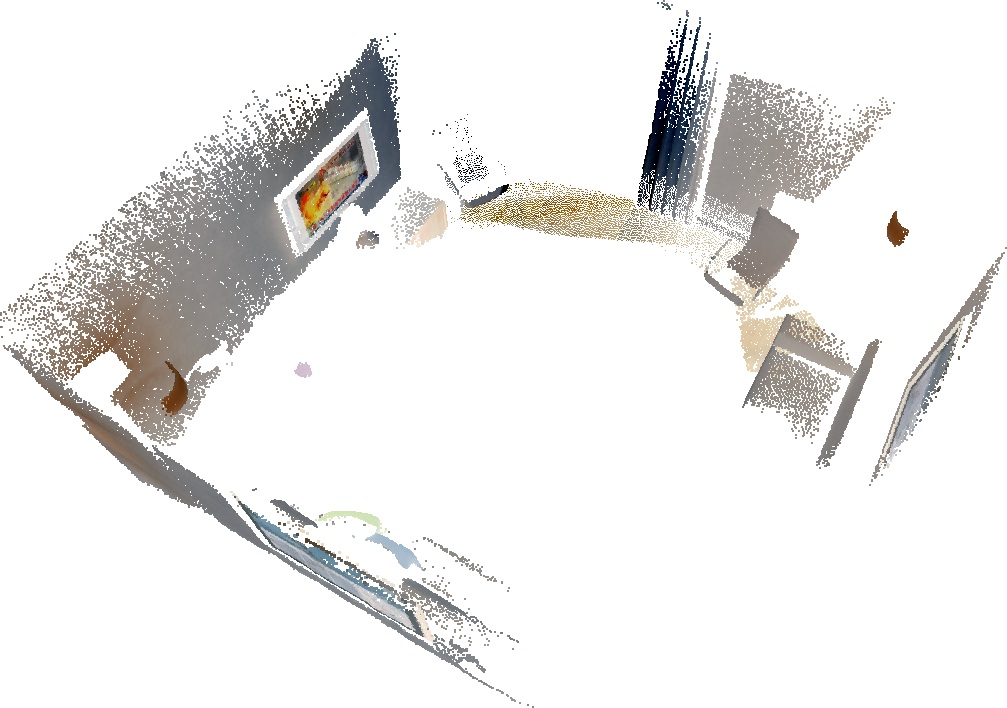}\caption{$\tau = 0.9$}\label{nid_q_1}\end{subfigure}
  \begin{subfigure}[t]{0.30\linewidth}\centering\includegraphics[width=4.5cm, height=3.5cm]{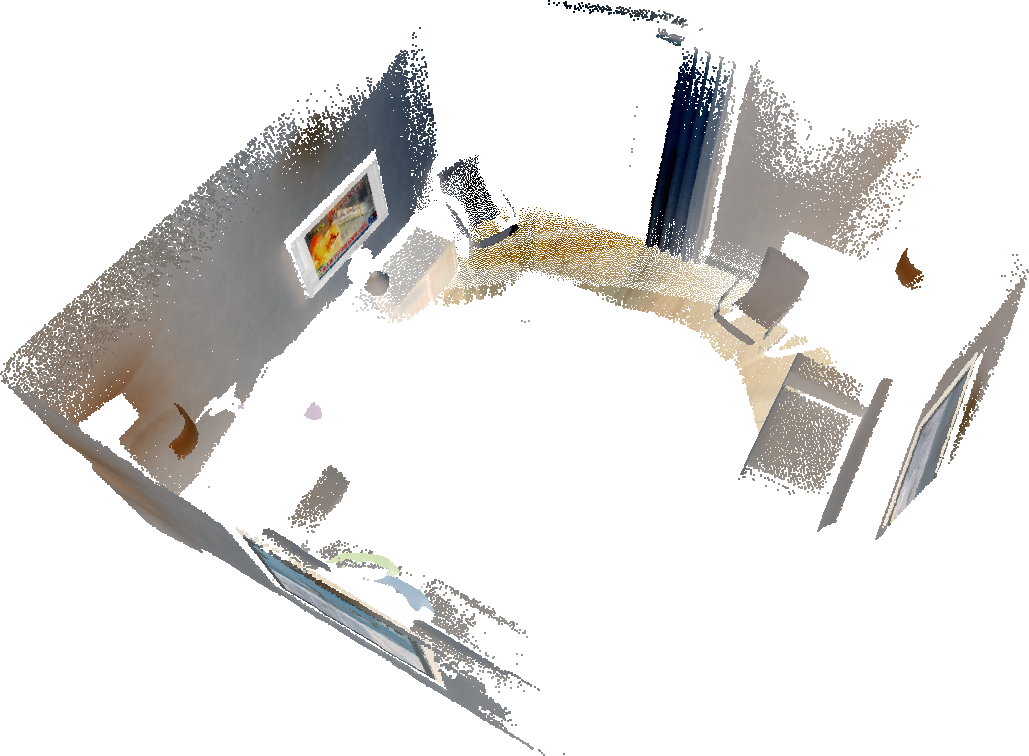}\caption{$\tau = 0.8$}\label{nid_q_3}\end{subfigure}
  \begin{subfigure}[t]
  {0.30\linewidth}\centering\includegraphics[width=4.5cm, height=3.5cm]{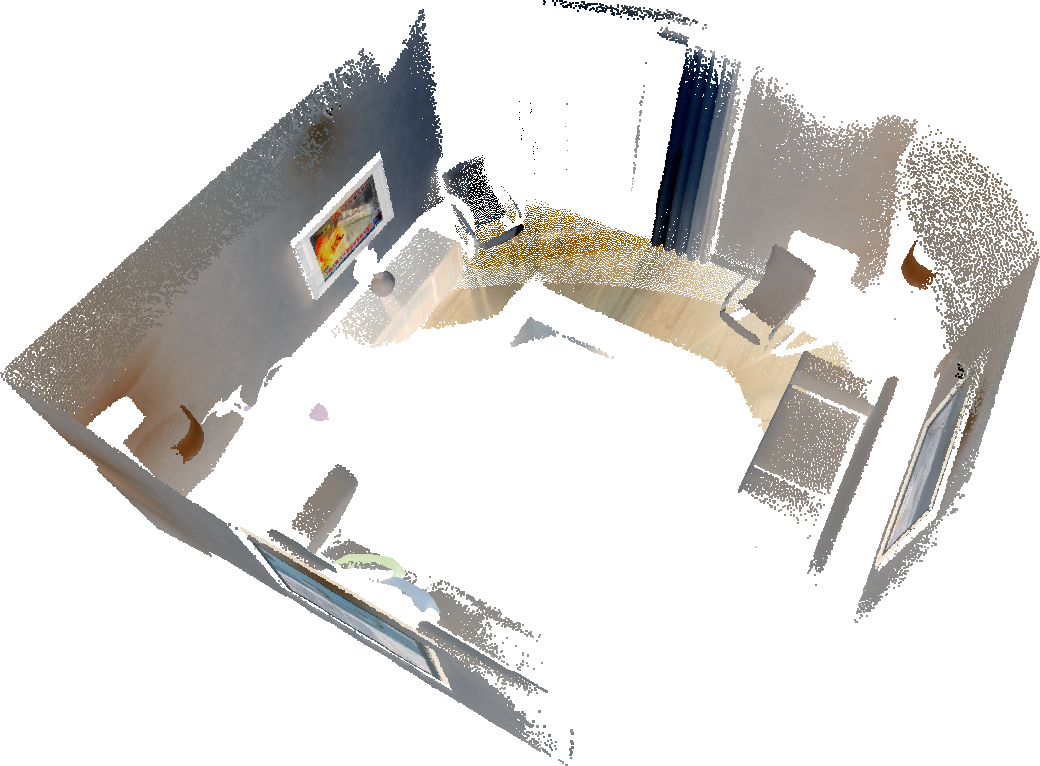}\caption{$\tau = 0.7$}\label{nid_q_5}\end{subfigure}
  \vspace{-1em}\caption{Qualitative results for the Kt1 sequence from the  ICL-NUIM dataset. For these experiments we set $\lambda_d = 0.75$.}
  \end{figure*}

   The EF map is divided into two subsets; $\theta_t$, containing all active surfels that have been fused within a window of time $\delta_t$, and $\phi_t$ containing all the inactive surfels. 
   Surfels in $\theta_t$ are used for tracking and fusion. As the camera loops back around into old regions of the map local loops are closed between $\theta_t$ and $\phi_t$
   and the surfels in $\phi_t$ that are viewed from the triggering frame are reactivated.\par
   To accommodate long sequences of frames without fusion it is important that we adjust $\delta_t$ in our system so that $\theta_t$ grows and shrinks and local loop closures are triggered in much the same way as in the original EF algorithm. This is achieved by letting $\delta_t = \delta_t + \delta_n$ where $\delta_n$ is the number of frames since the last frame that was fused. Additionally if a loop closure (local or global) is triggered then we proceed with fusion for that frame irrespective of the NID.\par 
   
   We take the current active frame, used for camera tracking and current inactive frame, used for closing local loops, as our reference frame for NID calculation. The depth maps for both frames are used to determine which values are visible from the current viewpoint. Doing so allows us to avoid a costly re-rendering of a single joint active/inactive frame, thereby making NID calculation a constant-time cost.\par

   \vspace{-0.55em}
\section{Experiments}\label{experiments}
\begin{figure}
  \centering
  \includegraphics[scale=0.35]{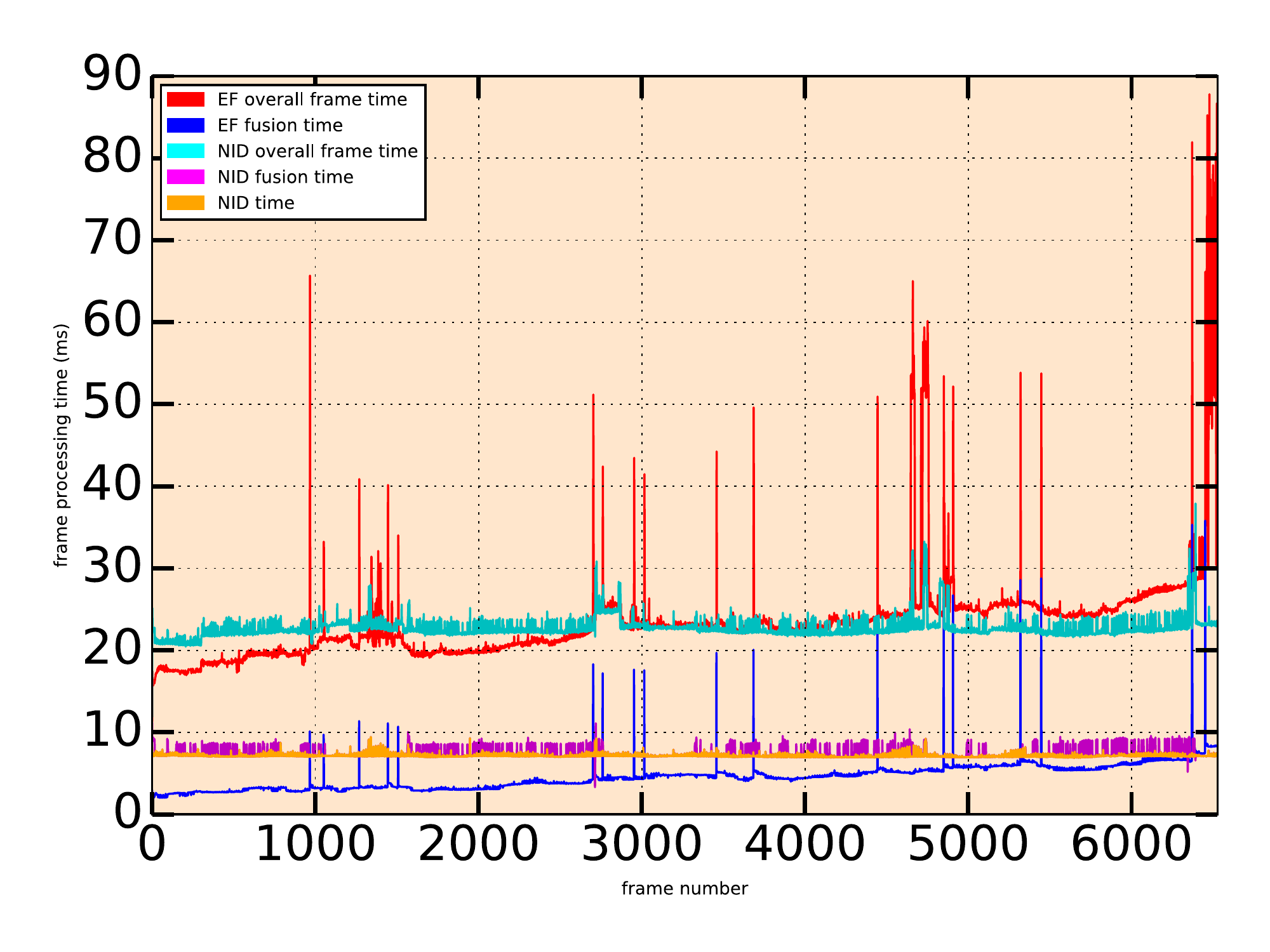}
  \vspace{-1.5em}
  \caption{Break down of frame processing time for the Dyson robotics lab sequence. While there is an overhead to computing the NID this cost is amortised over the course of the exploration of the office.}
  \label{dyson_lab_timings}
\end{figure}

\begin{figure*}[t]
\begin{subfigure}[t]{0.40\linewidth}\centering\includegraphics[scale=0.4]{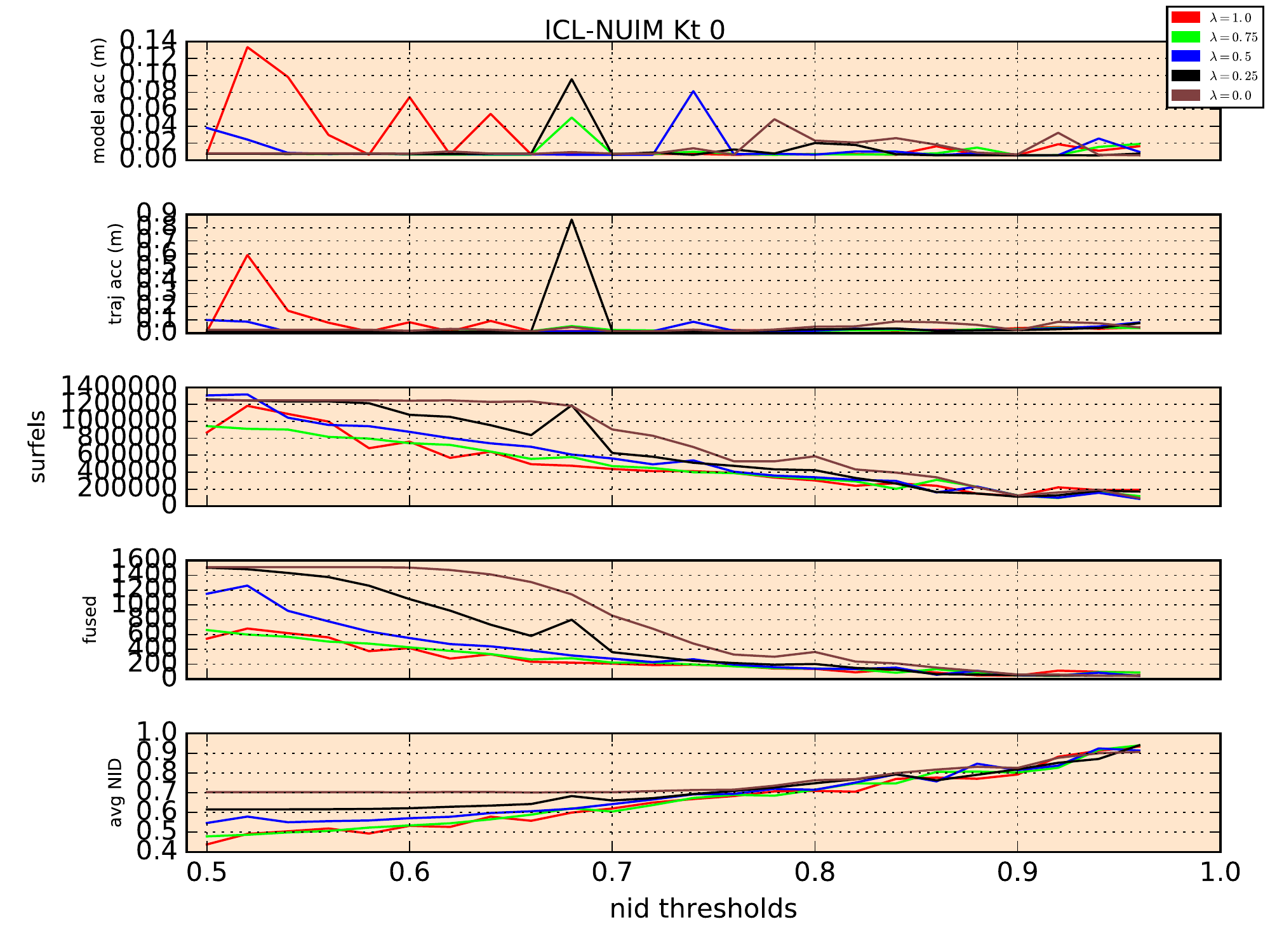}\label{kt0}\end{subfigure}
  \hspace{0.12\linewidth}
\begin{subfigure}[t]{0.40\linewidth}\centering\includegraphics[scale=0.4]{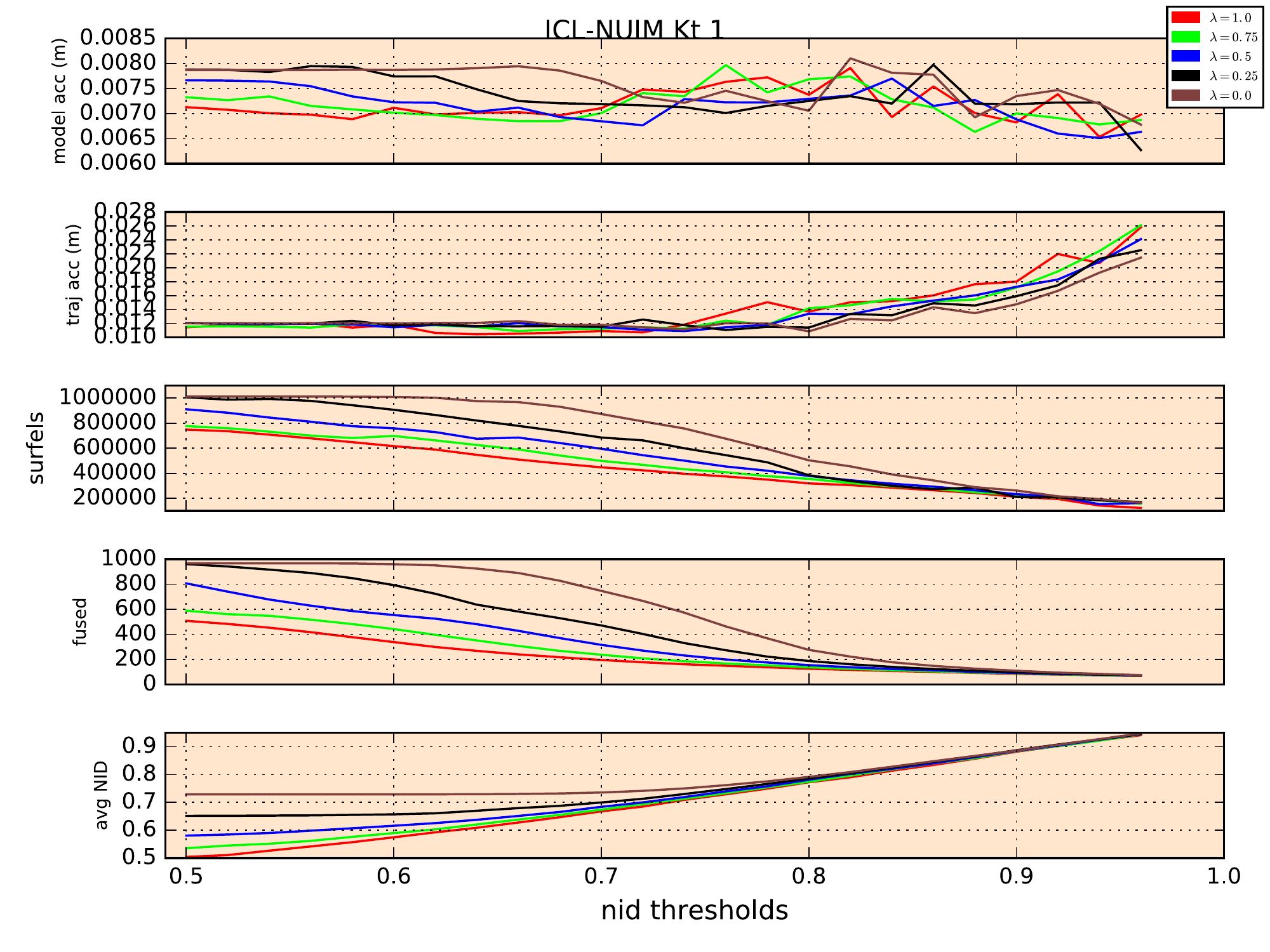}\label{kt1}\end{subfigure}

\vspace{-1em}\begin{subfigure}[t]{0.40\linewidth}\centering\includegraphics[scale=0.4]{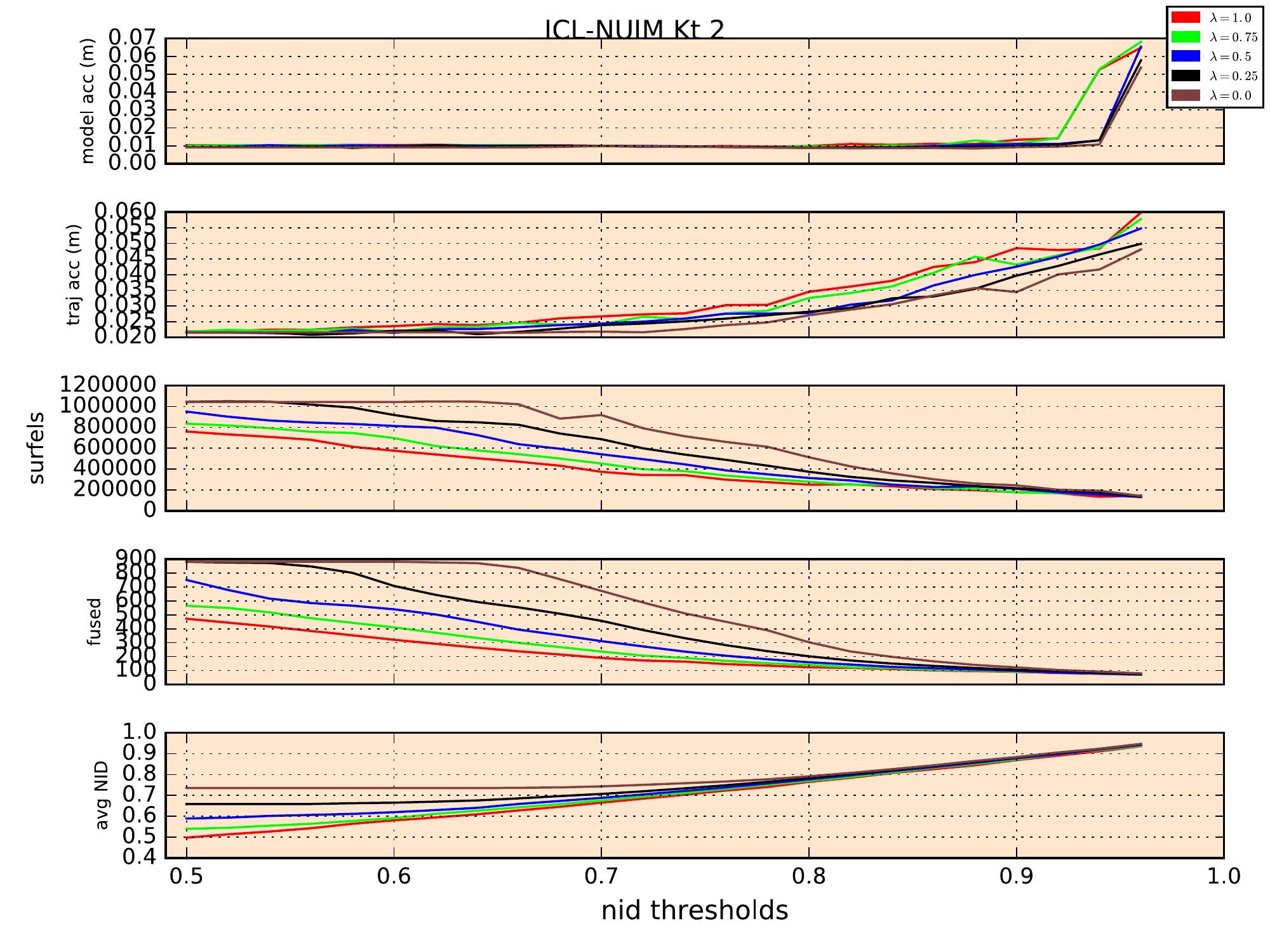}\label{kt2}\end{subfigure}
  \hspace{0.12\linewidth}
\begin{subfigure}[t]{0.40\linewidth}\centering\includegraphics[scale=0.4]{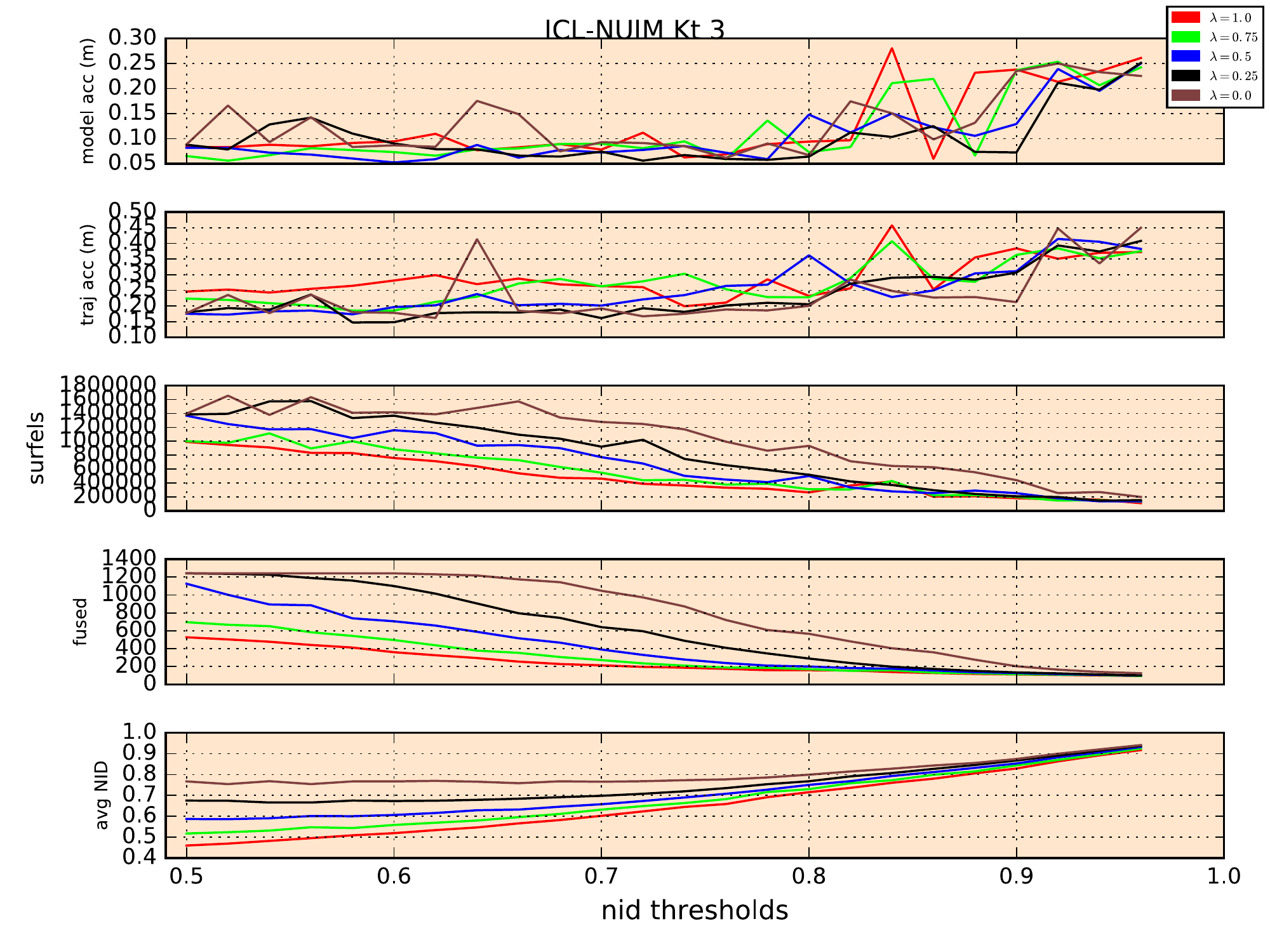}\label{kt3}\end{subfigure}
  \vspace{-1.75em}
\caption{Results for the ICL-NUIM sequences. We measured the model accuracy, trajectory accuracy, number of surfels in the final model, number of frames fused and the average NID score as we varied the the NID threshold and the relative weighting between $NID_d$ and $NID_{rgb}$.}
\label{results_icl}
\end{figure*}

In this section we provide experimental results of the application of the system to a number of  datasets. Figure~\ref{nid_example} shows different points during the reconstruction of Fri 3 Office sequence~\cite{freiburg}. Here, the effect of NID can be seen in reducing the number of frames fused whilst maintaining the accuracy and coverage of the reconstruction.

To provide a quantitative evaluation of the system's performance we use the ICL-NUIM synthetic living room dataset~\cite{iclnuim} where we measure the impact of varying both the NID threshold and the relative weighting factor, $\alpha$.The results are displayed in Figure~\ref{results_icl} across a number of metrics (see caption for more details). 
These results allow direct comparison to the original EF algorithm where again our system can maintain comparable accuracy whilst using significantly fewer frames and producing significantly fewer surfels in the final model.\par

Finally, to measure the scalability of our approach we used the Dyson lab sequence~\cite{elasticfusionjournal} which consists of over $6500$ frames captured during a scan of an office environment. The sequence contains a mix of segments where the camera explores new regions and where the camera closes local loops and re-visits already scanned regions of the office. These results are shown in Figure \ref{dyson_lab_timings}. Note that in this plot the NID fusion time includes the time for computing the NID, which, as we see, dominates.
Furthermore, whilst the EF fusion time increases steadily over the log, the rate of increase for the fusion component of the NID fusion approach is negligible. This is both due to the NID based frame selection, and the consequent reduction in  surfels. Hence whilst we do not see a substantial difference in the overall frame processing times of the two approaches over the time scale of this log, it can be seen that the rates of growth are very different. Our current focus is on using an image pyramid to reduce the NID computation time and achieve a more pronounced benefit from the NID based approach.\par 
   \vspace{-1em}
   \section{Conclusion}\label{conclusion}
   We have presented a frame sampling strategy based on NID that avoids fusing redundant frames in an EF based 3D dense reconstruction pipeline thereby using far fewer frames to reconstruct models of comparable accuracy to the original EF algorithm.
   In future work we aim to perform a full comparative analysis of NID sampling against other frame sampling methods such as uniform sub-sampling and motion thresholding. We also aim to incorporate the technique into a   dense collaborative mapping platform, such as ~\cite{collabef}, to significantly increase the scalability of the approach. \par
   
   {\small
   \bibliographystyle{ieee_fullname}
   \bibliography{conversionreferences}
   }

\end{document}